\def\year{2017}\relax
\documentclass[letterpaper]{article}
\usepackage{aaai17}
\usepackage{times}
\usepackage{helvet}
\usepackage{courier}

\usepackage{amsmath}

\usepackage[pdftex]{graphicx}
\graphicspath{{images/}}

\usepackage{array}
\newcolumntype{L}[1]{>{\raggedright\let\newline\\\arraybackslash\hspace{0pt}}m{#1}}
\newcolumntype{C}[1]{>{\centering\let\newline\\\arraybackslash\hspace{0pt}}m{#1}}
\newcolumntype{R}[1]{>{\raggedleft\let\newline\\\arraybackslash\hspace{0pt}}m{#1}}

\usepackage{multirow}
\usepackage{flushend}
\usepackage{paralist}

\begin{document}
%
\title{A Deep Multi-task Learning Approach to Skin Lesion Classification}
\author{
Haofu Liao, Jiebo Luo\\
Department of Computer Science, University of Rochester\\
Rochester, New York 14627, USA\\
Email: \{hliao6, jluo\}@cs.rochester.edu\\
}
\maketitle
\begin{abstract}
Skin lesion identification is a key step toward dermatological diagnosis. When
describing a skin lesion, it is very important to note its body site distribution
as many skin diseases commonly affect particular parts of the body. To exploit the
correlation between skin lesions and their body site distributions, in this
study, we investigate the possibility of improving skin lesion classification
 using the additional context information provided by body location. Specifically, we build
a deep multi-task learning (MTL) framework to jointly optimize skin lesion classification and 
body location classification (the latter is used as an inductive bias). 
Our MTL framework uses the state-of-the-art
ImageNet pretrained model with specialized loss functions for the two related tasks. Our
experiments show that the proposed MTL based method performs more robustly
than its standalone (single-task) counterpart.

\end{abstract}

\section{Introduction} \label{sec: intro}

Visual aspects of skin diseases, especially skin lesions, play a key role in dermatological
diagnosis. A successful identification of the skin lesion allows skin disorders
to be placed in certain diagnostic categories where specific diagnosis can be established
 \cite{cecil2012goldman}. However, categorization of skin lesions is a challenging process. It usually involves identifying the specific morphology,
distribution, color, shape and arrangement of lesions. When these components are
analyzed separately, the differentiation of skin lesions can be quite complex and
requires a great deal of experience and expertise \cite{lawrence2002physical}.

Besides the high requirement of expertise, the categorization of skin lesions by human
is essentially subjective and not always reproducible. To achieve a more objective and
reliable lesion recognition and ease the process of dermatological diagnosis, a
computer-aided skin lesion classification system should be considered.
Traditional approaches to computer-aided skin lesion/disease classification usually
focus on certain types of skin diseases, such as melanoma and basal cell carcinoma,
where the visual aspects of skin lesions are more regular and predictable. In those
cases, human-engineered feature extraction algorithms can be easily developed.
However, when we extend the lesion types to a broader range where all the possible
combinations of lesional characteristics need to be considered, human-engineered feature
extraction becomes infeasible and the traditional approaches fail.

Deep convolutional neural networks (CNNs) have shown to be very successful in recent years.
Specifically, the vision challenges from ILSVRC \cite{DBLP:journals/ijcv/RussakovskyDSKS15}
and MS COCO \cite{DBLP:conf/eccv/LinMBHPRDZ14} show that contemporary CNN
architectures are able to surpass human in many vision tasks. One thing behind
the success of CNN is its ability to do feature engineering automatically
from a large-scale dataset. It has been reported by many studies \cite{DBLP:conf/cvpr/RazavianASC14,DBLP:conf/icml/DonahueJVHZTD14,DBLP:conf/eccv/ZeilerF14,liao2016deep} that features extracted by contemporary
CNNs yield consistent superior results compared to the highly tuned non-CNN counterparts
in many tasks. Therefore, in this study, we propose to develop a skin lesion
classification model based on the state-of-the-art CNN architectures.

However, instead of treating the skin lesion classification as a standalone problem
and training a CNN model using skin lesion labels only, we further propose to
jointly optimize the skin lesion classification with a related auxiliary task,
body location classification. The motivation behind this design is to make use of the
body site predilection of skin diseases \cite{cox2004diagnosis}
as it has long been recognized by dermatologists that many skin diseases and their corresponding
skin lesions are correlated with their body site manifestation. For example, a skin
lesion caused by sun exposure is only present in sun-exposed areas of the body
(face, neck, hands, arms) \cite{cecil2012goldman}. Therefore, a CNN architecture
that can exploit the domain-specific information contained in the body locations
should be intuitively helpful in improving the performance of our skin lesion classification
model.

In this study, we present a multi-task learning framework for universal skin lesion (all lesion types)
classification using deep convolutional neural networks. In order to learn a
wide variety of visual aspect of skin lesions, we first collect 21657 images from
DermQuest (www.dermquest.com), a public skin disease atlas contributed
by dermatologists around the world. We then formulate our model into a dual-task
based learning problem with specialized loss functions for each task. Next, to
boost the performance, we fit our model into the state-of-the-art deep residual
network (ResNet) \cite{DBLP:journals/corr/HeZRS15}
which is the winning entry of ILSVRC 2015 \cite{DBLP:journals/ijcv/RussakovskyDSKS15} and
MS COCO 2015 \cite{DBLP:conf/eccv/LinMBHPRDZ14}.

\noindent \textbf{Contribution:} To our best knowledge, this is the first attempt to
target the universal skin lesion classification problem systematically using a deep
multi-task learning framework. We show that the jointly learned representations from
body locations indeed facilitate the learning for skin lesion classification. Using
the state-of-the-art CNN architecture and combining the results from different
models we can achieve as high as a 0.80 mean average precision (mAP) in classifying
skin lesions.

\section{Related Work}

Most of the existing works \cite{arroyo2014automated,xie2014dermoscopy,Fabbrocini:2014yq}
only focus on one or a few skin disease and solve the problem using conventional
machine learning approach, i.e., extracting manually engineered features from
segmented lesion patches and classifying with a linear classifier such as SVM.
While in our study, we target a more challenging problem where all skin diseases
are considered.

Many CNN related approaches have been proposed to solve dermatology problems in recent years.
Some works \cite{cruz2014automatic,wang2014cascaded,arevalo2015unsupervised} used
CNNs as an unsupervised feature extractor and detect mitosis, an indicator of cancer,
from histopathology images. \cite{esteva2015deep} presented a CNN architecture
for diagnosis-targeted skin disease classification. They trained their model
with a contemporary CNN architecture using a large-scale dataset (23000 images). Similar
to our study, they also tried to classify skin diseases in a broader range.
What sets us apart from their work is instead of training with diagnosis labels
and making diagnostic decision directly, our work classifies skin diseases by their
lesional characteristics. According to a recent study \cite{liao2016skin}, skin
lesion is proven to be a more appropriate subject for skin disease classification as
many diagnoses can not be distinguished visually. Recently, \cite{DBLP:conf/isbi/KawaharaBH16}
also proposed a CNN based model to classify skin lesions for non-dermoscopic images. However,
they only managed to build their model on a prior art CNN architecture with a relatively small
dataset (1300 images).

Multi-task learning (MTL) \cite{DBLP:journals/ml/Caruana97} is an approach to
learning a main task together with other related tasks in parallel with the goal 
of a better generalization performance. Learning multiple tasks jointly has been
proven to be very effective in many computer vision problems, such as attribute
classification \cite{DBLP:journals/corr/HandC16}, face detection \cite{DBLP:journals/corr/RanjanPC16},
face alignment \cite{DBLP:journals/pami/ZhangLLT16} and object detection \cite{DBLP:conf/nips/RenHGS15}.
However, we find no multi-task learning based algorithm has been developed for
dermatology related problems.

\section{Dataset}

\begin{figure}
  \centering
  \includegraphics[scale=0.51]{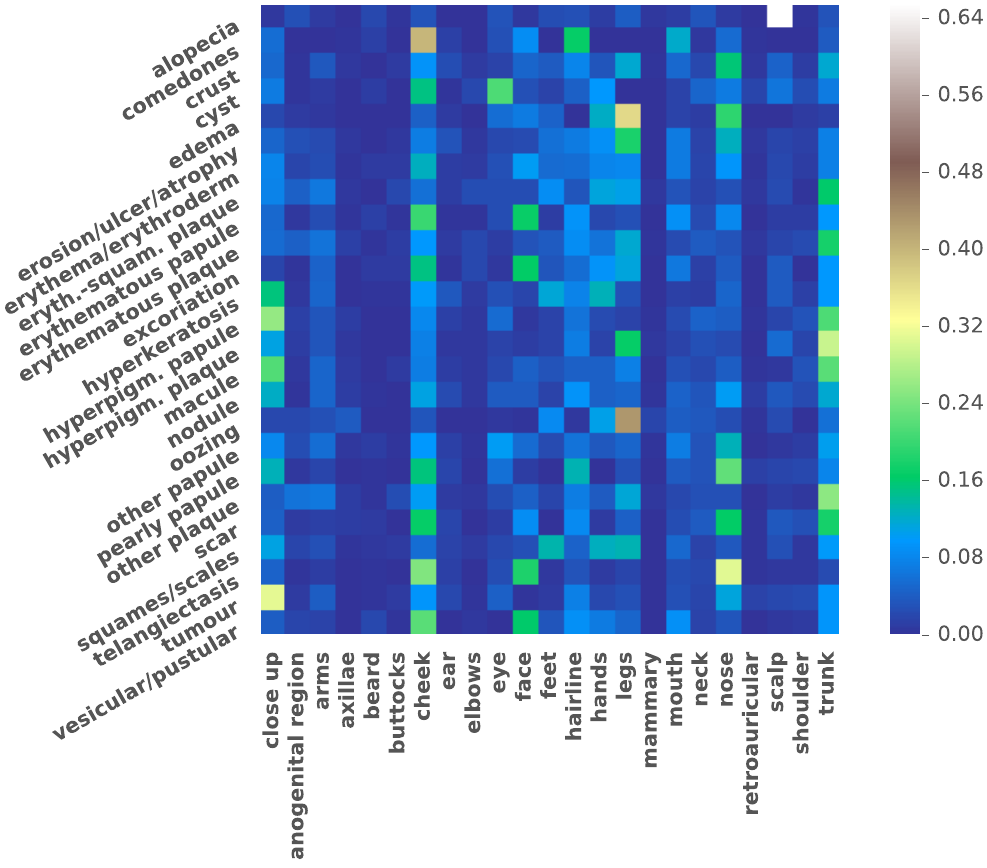}
  \caption{The correlation matrix between skin lesion and body location. Each row
  denotes a skin lesion and each column denotes a body location. A cell at
  $(i, j)$ denotes the proportion of the images with both label $i$ and 
  label $j$ among all the $i$ images (best viewed in color).}
  \label{fig: correlation}
\end{figure}

\begin{figure*}[!ht]
  \centering
  \includegraphics[scale=0.36]{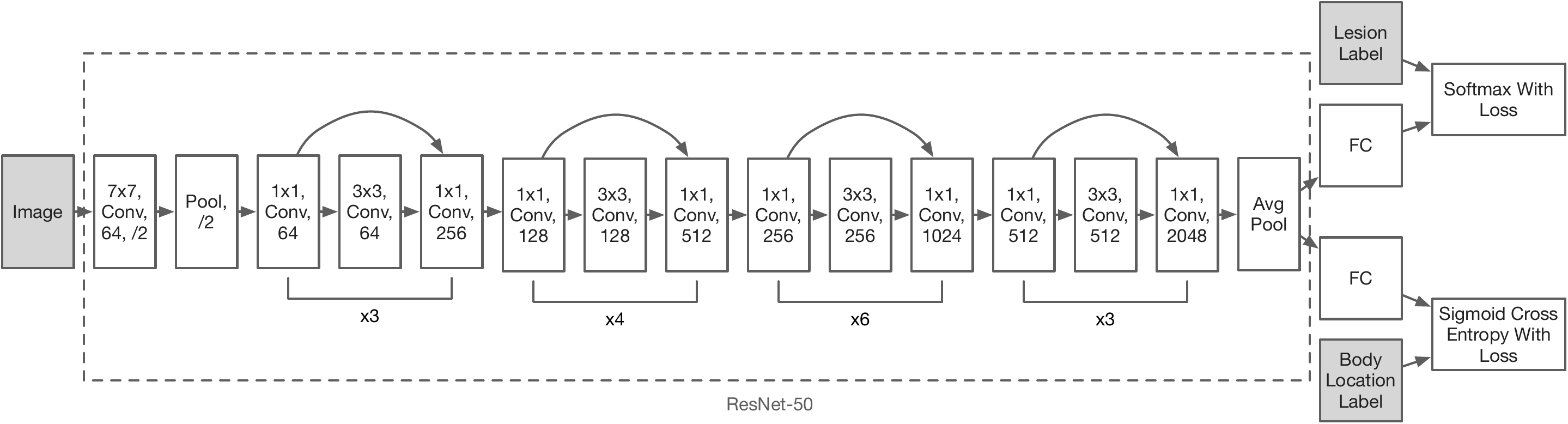}
  \caption{The network structure of the proposed method. ``Conv'' denotes the
  convolutional layer, ``Pool'' denotes the pooling layer and ``FC'' denotes the
  fully connected layer. The three dark blocks are the data layers for images,
  skin lesions, and body locations, respectively. The net architecture inside
  the dotted area is identical to the ResNet-50 network.
  }
  \label{fig: architecture}
\end{figure*}

All the dermatology images used in this study are collected from DermQuest. We choose
DermQuest against other dermatology atlantes is because it has the most detailed
annotations and descriptions for each of the dermatology image and it is the only
public dermatology atlas that contains both skin lesion and body location labels.
Most of the dermatology images from DermQuest are contributed by individual dermatologists.
When contributing an image, they are required to input the descriptions
(diagnosis, primary lesions, body location, pathophysiology, etc.) by their own.
As the terminology used by dermatologists are not unified, they may use different
terms and morphologies when describing a dermatology image which results in an
inconsistency of DermQuest images.

Due to the inconsistency, there are  180 lesion types in total in the DermQuest atlas
, which is larger than any of the existing lesion morphology. Therefore, with the
help of a dermatologist, we refined the list of lesion types to make sure they
reasonably and consistently represent the lesional characteristics of the images
in DermQuest. We refine and merge lesions based on the lesion morphology described in \cite{cox2004diagnosis}
with some modifications:
\begin{inparaenum}[1)]
  \item We removed those infrequent lesion types (less than $10$ images) as they do not
  have enough images for our model to learn some meaningful features.
  \item For some popular (greater than $1000$ images) sublesion types, such as hyperpigmented
  papule lesion under the papule family, we do not merge them as there are enough images in the 
  dataset so that our model can distinguish them from other sublesions under the
  same family.
  \item Some of the lesion types have common visual characteristics, such atrophy, erosion and
  ulcer, we also merge them together.
\end{inparaenum}
After the refinement, we finally come up with a lesion morphology list with $25$
lesions types for the DermQuest images. Note that there might be multiple 
lesion labels associated with an image as a skin disease usually manifests
different lesional characteristics at a time.

For the body location labels, the terminology used is more consistent. We do not
modify too much except we removed those infrequent labels as we did for the lesions. We also
merged some body locations that are too specific to not be mixed with its nearby
regions in an image. For example, an image labeled with nails usually contains
parts of the fingers. Thus, it is actually hard to tell whether it should be labeled
with nails or fingers. Hence, we directly merge them into the ``hands'' category.
There are $23$ body locations in the final list.

We also investigate the correlation between skin lesions and body locations
among images in DermQuest. The correlation map is shown in Figure \ref{fig: correlation}.
Here, each row denotes a skin lesion and each column denotes a body location.
Let $N_i$ denote the total number of images in our dataset that has lesion $i$
and $M_j$ denote the total number of images that has body location $j$. Then, the
cell at $(i, j)$ can be computed by
\begin{equation}
R_{ij} = \frac{N_i \cap M_j}{N_i}
\end{equation}
Thus, if a skin lesion frequently appears on certain body location, we will see
a high very value of $R_{ij}$. Notice that we have $23$ body location types. Thus,
if a skin lesion has no specific predilection of body locations, then cells in
the corresponding row should all have values close to $1/23$, i.e., dark blue in
the color bar. For example, the cells in row ``erythema/erythroderm'' are almost
in blue, which means ``erythema/erythroderm'' has little body location predilection.
This is consistent with our knowledge that ``erythema/erythroderm'' is a very
commonly seen lesion that can exists anywhere in the body. We can also see that 
``alopecia'' is highly correlated with ``scale''. It makes sense as ``alopecia''
is a lesion that related with hair loss.

\section{Methodology}

\subsection{Deep Multi-task Learning}

To jointly optimize the main (skin lesion classification) and auxiliary (body
location classification) tasks, we formulate our problem as follows. Let
$(\mathbf{X}_i, \mathbf{u}_i, v_i), i \in \{1, \dots N\}$ denotes the $i$th
data in the training set, where $\mathbf{X}_i$ is
the $i$th image and $\mathbf{u}_i$ and $v_i \in \{1, \dots, Q\}$ are the $i$th
labels for the skin lesion and body location, respectively. As multiple lesion
types may be associated with a dermatology image, the lesion label
$\mathbf{u}_i = [u_1^i, u_2^i, \dots, u_P^i]$ is a binary vector with 
\begin{equation}
  u_j^i = \begin{cases}
    1, & \text{if the $j$th lesion is associated with $\mathbf{X}_i$,} \\
    0, & \text{otherwise.}
  \end{cases}
\end{equation}
Here, $P$ and $Q$ denotes the number of skin lesions and body locations in our
dataset. Our goal is to minimize the objective function
\begin{align}
\mathcal{L} (\mathbf{W}) = 
& \frac{1}{N}\sum_{i=1}^N\ell_{les}(\mathbf{X}_i, \mathbf{u}_i; \mathbf{W}) + \nonumber \\
& \frac{1}{N}\sum_{i=1}^N\ell_{loc}(\mathbf{X}_i, v_i; \mathbf{W}) + \Phi(\mathbf{W})
\end{align}
in which $\Phi(\cdot)$ is a regularization term, $\ell_{les}(\cdot)$ is the loss function
for skin lesions and $\ell_{loc}(\cdot)$ is the loss function for body locations. 

Since there might be multiple lesions associated with an input image, we use
a sigmoid cross-entropy function for the skin lesion loss so that each lesion
can be optimized independently. Let $s_j(\mathbf{X}_i; \mathbf{W}), j \in \{1, \dots, P\}$
denotes the $j$th output of the last fully-connected (FC) layer for the skin lesions.
Then the $j$th activation of the sigmoid layer can be written as
\begin{equation}
a_j(\mathbf{X}_i; \mathbf{W}) = \frac{1}{1 + e^{-s_j(\mathbf{X}_i; \mathbf{W})}}.
\end{equation}
and the corresponding cross-entropy loss is
\begin{align}
\ell_{les}(\mathbf{X}_i, \mathbf{u}_i; \mathbf{W}) = & - \sum_{j=1}^Pu_j^i\log{a_j(\mathbf{X}_i; \mathbf{W})} + \nonumber \\
& (1-u_j^i)\log{(1 - a_j(\mathbf{X}_i; \mathbf{W}))}.
\end{align}
For the body locations, it is a many-one classification problem. Thus, we use
a softmax loss function so that only one label will be optimized each time.
Let $t_j(\mathbf{X}_i; \mathbf{W}), j \in \{1, \dots, Q\}$ denotes the $j$th output
of the last FC layer for the body locations.
Then the $j$th activation of the softmax layer can be written as
\begin{equation}
b_j(\mathbf{X}_i; \mathbf{W}) = \frac{e^{t_j(\mathbf{X}_i; \mathbf{W})}}{\sum_ke^{t_k(\mathbf{X}_i; \mathbf{W})}}
\end{equation}
and the corresponding softmax loss is
\begin{equation}
\ell_{loc}(\mathbf{X}_i, v_i; \mathbf{W}) = -\log(b_{v_i}(\mathbf{X}_i; \mathbf{W}))
\end{equation}
Finally, for the regularization term, we use the L2 norm
\begin{equation}
\Phi(\mathbf{W}) = \gamma\|\mathbf{W}\|_2
\end{equation}
where the regularization
parameter $\gamma$ controls the trade off between the regularization term and
the loss functions.

\subsection{Implementation}

The architecture of the proposed method is given in Figure \ref{fig: architecture}.
We build our CNN architecture on top of ResNet-50 (50 layers). Though it is
possible to use a deeper ResNet to get a marginal performance gain, ResNet-50 is
considered sufficient for this proof-of-concept study. To facilitate our goal in MTL,
three data layers are used. One data layer is for the images and the other two data layers
are for the lesion labels and body location labels, respectively. We then remove
the finally FC layer in the original ResNet and add two sibling FC layers, one
for the skin lesions and the other for the body locations. After the FC layers,
we add a sigmoid cross entropy loss layer for the skin lesion classification and
a softmax layer for the body location classification.

We use the Caffe deep learning framework \cite{DBLP:conf/mm/JiaSDKLGGD14} for
all of our experiments and run the programs with a GeForce GTX 1070 GPU. As transfer
learning has shown to be more effective in image classification problems
\cite{DBLP:conf/cvpr/RazavianASC14}, instead of training from scratch, we
initialize our network from the ImageNet \cite{DBLP:conf/cvpr/DengDSLL009} pretrained
ResNet-50 model \footnote{We also trained the network from scratch but no
performance gain was observed.}. As a dermatology image may be taken from different
distances, the scale of certain skin lesions may vary. Thus, following the practice in \cite{DBLP:journals/corr/SimonyanZ14a}, we scale each image with its shorter side length randomly selected
from $[256, 480]$. This process is called scale jittering. Then we follow the ImageNet
practice in which a 224 x 224 crop is randomly sampled from the mean subtracted
images or their horizontal flips. In the testing phase, we perform the standard
10-crops testing using the strategy from \cite{DBLP:conf/nips/KrizhevskySH12}. 

For the hyper-parameters, we use SGD with a mini-batch size of $20$ and set the
momentum to $0.9$ and the weight decay (the regularization parameter)
to $0.0001$. The initial learning rate is $0.001$ and is reduced by $0.1$ when
error plateaus. It is worth mentioning that the two newly added FC layers have
bigger learning rate multipliers ($10$ for the weights and $20$ for the bias) so
that their effective learning rate is actually $0.01$. We use higher learning rate 
for these two layers is because their weights are randomly initialized. The model
is trained for up to $12 \times 10^4$ iterations. Note that this is a relatively
large number for fine-tuning. This is because the scale jittering greatly augmented our
dataset and it takes longer time for the training to converge. During the training,
we do not see any over-fitting from the validation set.

\section{Experimental Results}

In this section, we investigate the performance of the proposed method on both the
skin lesion classification and body location classification tasks. In all of our
experiments, we use data collected from DermQuest. In total, there are $21657$
images that contain both the skin lesion and body location labels. To avoid
overfitting, 5-folds cross-validation is used for each experiment.

\subsection{Performance of Skin Lesion Classification}

For skin lesion classification, since it is a multi-label classification problem,
we use mean average precision (mAP) as the evaluation metrics following the practice
in VOC \cite{DBLP:journals/ijcv/EveringhamGWWZ10} and ILSVRC. In this study, we
use two different mAPs: 
\begin{inparaenum}[1)]
  \item class-wised mAP, where we treat the sorted evaluations of all images on
  certain class as a ranking and compute the mAP over the classes.
  \item image-wised mAP, where we treat the sorted evaluations of all classes on
  certain image as a ranking and compute the mAP over the images.
\end{inparaenum}
Formally put, the two metrics can be computed using the following formulas:
\begin{equation}
  \text{mAP-class} = \frac{1}{P}\sum_{i=1}^P\sum_{j=1}^Np_i(j)\Delta r_i(j),
\end{equation}
\begin{equation}
  \text{mAP-image} = \frac{1}{N}\sum_{i=1}^N\sum_{j=1}^Pq_i(j)\Delta s_i(j),
\end{equation}
Here, $N$ is the total number of images, $P$ is the total number of classes,
$p_i(j)$ is the precision of the ranking for class $i$ at cut-off $j$ and
$\Delta r_i(j)$ is the difference of the recall (of the ranking for class $i$) from
cut-off $j-1$ to $j$. $q_i(j)$ and $\Delta s_i(j)$ can be defined similarly to
$p_i(j)$ and $\Delta r_i(j)$. 

We compare our proposed method with two standalone architectures (single task) based
on AlexNet and ResNet-50, respectively. For the hyper-parameters of AlexNet, we use the settings
from \cite{DBLP:conf/nips/KrizhevskySH12}, i.e., batch size $= 256$, momentum $= 0.9$
and weight decay $= 0.0005$. For the standalone ResNet-50, we use the same hyper-parameter
settings as our proposed method. Both the two architectures are fine-tuned from ImageNet
pretrained models with learning rate set to $0.01$.

\begin{figure*}[!ht]
  \centering
  \includegraphics[scale=0.45]{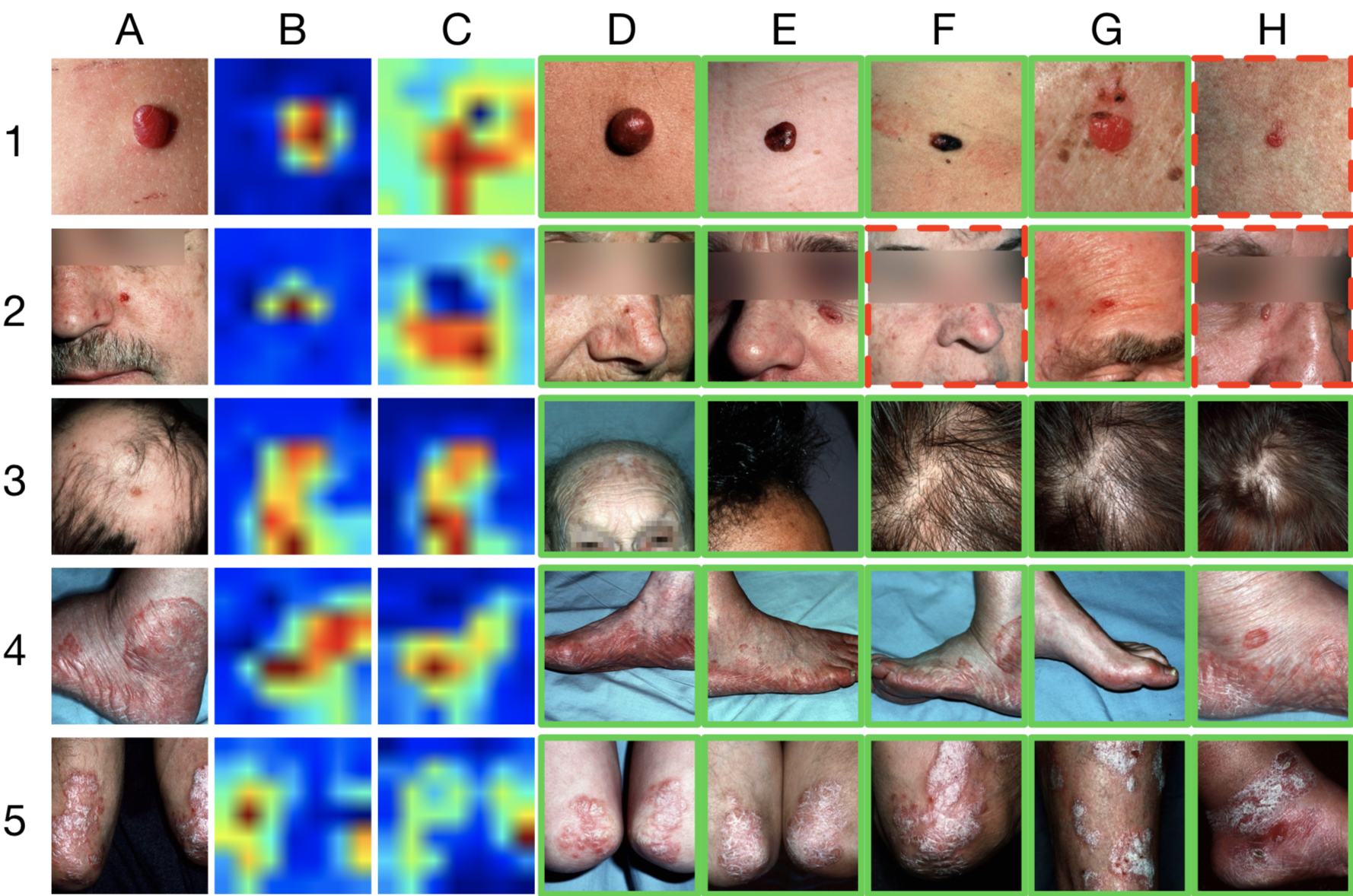}
  \caption{Image retrieval. A: the query images. B: image attention of the primary lesion.
  C: image attention of the body location. D-H: retrieved images. Retrieved image
  with red dotted border means it has no common lesion labels with the query image.
  Primary lesions of 1-5: nodule, erosion/ulcer/atrophy, alopecia, erythematous
  squamous plaque, squames/scales.
  }
  \label{fig: retrival}
\end{figure*}

\begin{table}[!ht]
  \setlength{\tabcolsep}{2pt}
  \centering
  \resizebox{\linewidth}{!}{
    \begin{tabular}{c|C{1.4cm}|C{1.4cm}|C{1.4cm}|C{1.4cm}}
    \hline \hline
    \multirow{2}{*}{Lesion Type} & \multicolumn{4}{c}{Average Precision} \\
    \cline{2-5}
    & AlexNet & ResNet & MTL & Ensemble \\ 

    \hline
    alopecia & 0.763 & 0.845 & 0.843 & \textbf{0.855} \\
    comedones & 0.687 & 0.817 & \textbf{0.861} & 0.858 \\
    crust & 0.677 & 0.783 & 0.794 & \textbf{0.807} \\
    cyst & 0.461 & 0.625 & 0.698 & \textbf{0.702} \\
    edema & 0.633 & 0.707 & 0.751 & \textbf{0.758} \\
    \footnotesize{erosion/ulcer/atrophy} & 0.774 & 0.850 & 0.867 & \textbf{0.873} \\
    \footnotesize{erythema/erythroderm} & 0.742 & 0.820 & \textbf{0.844} & 0.843 \\
    \footnotesize{eryth.-squam. plaque} & 0.496 & 0.658 & 0.683 & \textbf{0.690} \\
    \footnotesize{erythematous papule} & 0.767 & 0.846 & 0.857 & \textbf{0.861} \\
    \footnotesize{erythematous plaque} & 0.538 & 0.670 & 0.704 & \textbf{0.708} \\
    excoriation & 0.467 & 0.605 & 0.635 & \textbf{0.651} \\
    hyperkeratosis & 0.643 & 0.772 & 0.796 & \textbf{0.802} \\
    \footnotesize{hyperpig. papule} & 0.589 & 0.690 & \textbf{0.738} & 0.730 \\
    \footnotesize{hyperpig. plaque} & 0.473 & 0.637 & \textbf{0.675} & \textbf{0.675} \\
    macule & 0.619 & 0.742 & \textbf{0.780} & 0.777 \\
    nodule & 0.704 & 0.793 & 0.813 & \textbf{0.820} \\
    oozing & 0.497 & 0.595 & \textbf{0.674} & 0.663 \\
    other papule & 0.344 & 0.559 & 0.600 & \textbf{0.603} \\
    pearly papule & 0.716 & 0.849 & 0.875 & \textbf{0.879} \\
    other plaque & 0.331 & 0.553 & 0.549 & \textbf{0.562} \\
    scar & 0.521 & 0.690 & \textbf{0.728} & 0.726 \\
    squames/scales & 0.591 & 0.704 & \textbf{0.748} & 0.746 \\
    telangiectasis & 0.655 & 0.821 & 0.837 & \textbf{0.848} \\
    tumour & 0.598 & 0.728 & 0.768 & \textbf{0.770} \\
    \footnotesize{vesicular/pustular} & 0.664 & 0.792 & 0.814 & \textbf{0.823} \\
    \hline
    mAP-class & 0.598 & 0.726 & 0.757 & \textbf{0.761} \\
    mAP-image & 0.704 & 0.778 & 0.792 & \textbf{0.798} \\
    \end{tabular}
  }
  \caption{Skin lesion classification results. ``AlexNet'' and
  ``ResNet'' are trained using skin lesion labels only. ``MTL'' is the proposed method. An
  ensemble of ``ResNet'' and ``MTL'' is given under ``Ensemble''.}
  \label{tab: lesion performance}
\end{table}

The classification results are shown in Table \ref{tab: lesion performance}. Here,
``AlexNet'' and ``ReNet'' are the two standalone architectures, ``MTL'' is our proposed method,
and ``Ensemble'' contains the ensemble results of ``ResNet'' and ``MTL''.
First, we can see ``ResNet'' outperforms ``AlexNet'' by a big leap
which shows that the use of the state-of-the-art CNN architecture helps a lot in
boosting the performance. Then, we also observe a decent performance improvement against ``ResNet''
when using our proposed method. It means the joint optimization with body
location classification can really benefit the learning of the lesional characteristics.
Finally, we find that the highest mAP can be achieved with an ensemble of
``ResNet'' and ``MTL'', i.e., choosing the best evaluation scores of the two
models for each image.

We further analyze the performance difference of each class between ``ResNet'' and
``MTL''. We find that, in general, if a skin lesion has a strong correlation with a body location, it will also
have a better performance gain when using ``MTL''. Typical examples are ``comedone'', ``edema'',
``hyperpigmented papule'', ``oozing'', and ``tumor''. They all have a strong correlation
with certain body locations and we see they also have at least a 4\% improvement when using ``MTL''.
However, there are some exceptions. For example, we do not see any improvement
from ``alopecia'' even though it has a very strong correlation with ``scalp''.
One possible reason is that the strong correlation between ``alopecia'' and ``scalp''
makes ``scalp'' bias too much to ``alopecia'' such that some variations won't be
learned. We will further verify this hypothesis in the later discussion.

\subsection{Image Retrieval and Image Attention}

Figure \ref{fig: retrival} shows the image retrieval and attention of the
proposed method. For image retrieval, we take the output of the last pooling layer (pool5)
of the ResNet as the feature vector. For each query image from the test set, we compare its features with all the images in the training set and outputs the 5-nearest neighbors (in euclidean distance)
as the retrieval. If a retrieved image matches at least one label of
the query image, we annotate it with a green solid frame. Otherwise, we annotate
it with a red dotted frame. We can see that the retrieved images are visually very similar
to the query image.

For image attention, we adapt the method in \cite{DBLP:journals/corr/ZhouKLOT15}.
We first fetch the output of the final convolution layer (res5c) and get a set of $2048$ $7\times7$ activation maps.
Next, we calculate the weighted average of the activation maps using the learned weights
from the final FC layer. As the weights of an FC layer is a $K \times 2048$ matrix
where $K$ is the number of outputs of the FC layer, we will get $K$ attention spots.
We select the attention map that corresponding to the ground truth of the input
image as the final image attention. As there are two FC layers in our architecture,
we obtain two attention maps (one for the skin lesion and the other for the
body location) for each input image.

In Figure \ref{fig: retrival}, Column B contains the image attention for the
primary skin lesions and Column C contains the image attention for the body
locations. In general, the image attention for skin lesions should focus more
on the lesion area and the image attention for body locations should focus more
on the body parts. For Row 1-2 and Row 4-5, it is almost the case and we can see
our trained model knows where it should pay attention to. However, for Row 3 (``alopecia''),
the skin lesion attention map and the body location attention map look very similar. It
means for a ``scalp'' image, the skin lesion classifier and the body location
classifier are trained to make similar decisions. That is when the skin lesion classifier
sees an image with scalp, it will almost always output an ``alopecia'' label. This
is too biased and it explained why we did not see a performance boost for the ``alopecia''
label.

\subsection{Performance of Body Location Classification}

\begin{figure}[!ht]
  \centering
  \includegraphics[scale=0.38]{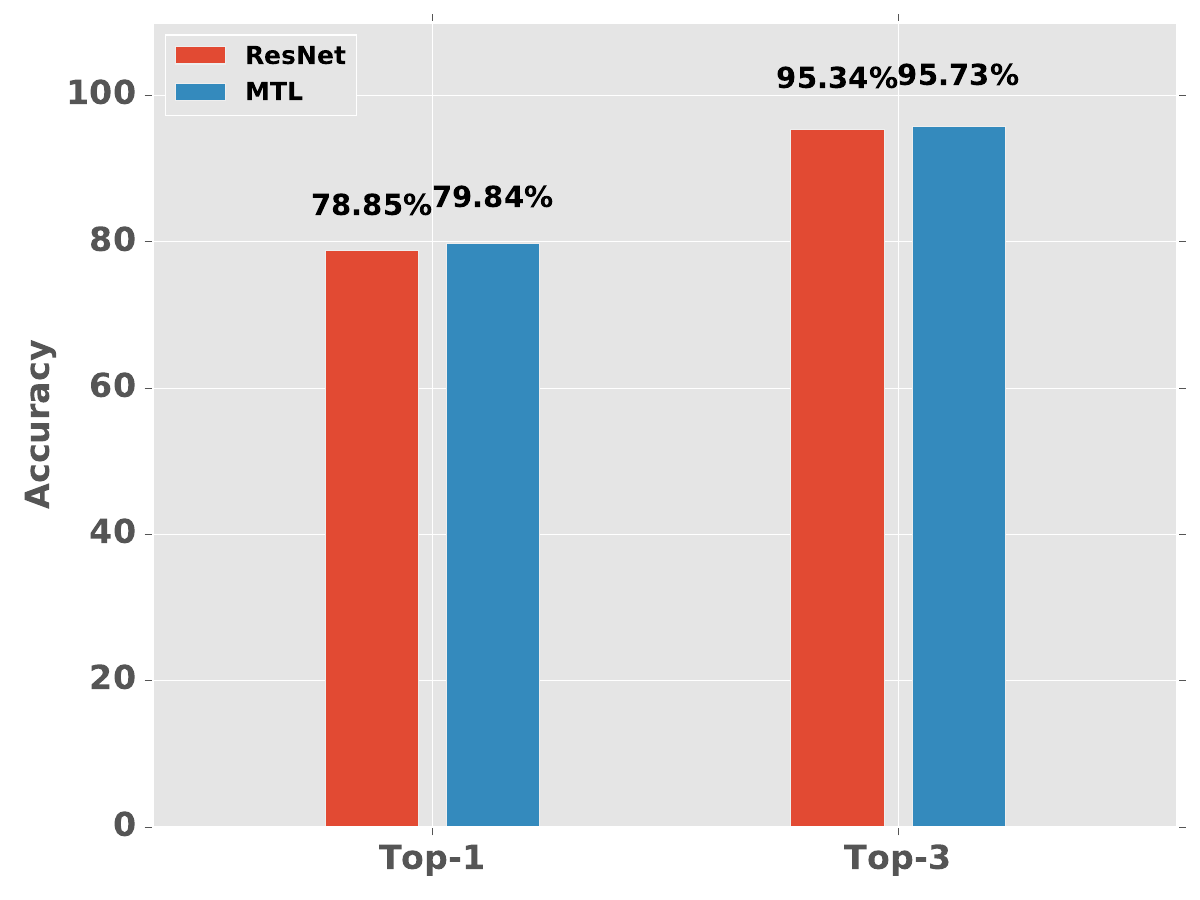}
  \caption{Body location classification results. ``ResNet'' is trained using body
  location only and ``MTL'' is the proposed multi-task learning method.}
  \label{fig: body performance}
\end{figure}

We also compare the performance of our method with its standalone counterpart
in classifying body locations. To this end, we fine-tune another ResNet-50 model
with body location labels only. For the evaluation metrics, the standard
top-$1$ and top-$3$ accuracies are used as body location classification is a
multi-class classification problem. The evaluation results are given in Figure
\ref{fig: body performance}. We can also see a performance improvement
from ``ResNet'' to ``MTL''. This is somewhat counter-intuitive as the classification
of a body location should have nothing to do with the skin lesions. However, as
we restrict the images to be dermatological images, a slight performance gain 
is reasonable.

\section{Conclusions}

We have developed a deep multi-task learning framework for universal skin
lesion classification. The proposed method learns
skin lesion classification and body location classification in parallel based
on the state-of-the-art CNN architecture. To be able to learn a wide variety of lesional characteristics
and classify all kinds of lesion types, we have also collected and built a large-scale
skin lesion dataset using images from DermQuest. The experimental results have shown that
\begin{inparaenum}[1)]
  \item Training using the state-of-the art CNN architecture on a large scale of skin
  lesion dataset leads to a universal skin lesion classification system with good performance.
  \item It is indeed beneficial to use the body location
  classification as an auxiliary task and train a deep multi-task learning based model
  to achieve improved skin lesion classification.
  \item An ensemble of the proposed method and its standalone counterpart can achieve an image-wise
  mAP as high as 0.80.
  \item The performance of body location classification is also improved under the deep multi-task learning framework.
  \item It is also confirmed by the obtained image retrieval and attention that the trained model not only learns the lesional features very well but also knows generally where to pay attention to.
\end{inparaenum}
Our future work includes integrating the image analysis with other patient information to build an overall high-performance diagnosis system for diseases with skin lesion symptoms. 

\section{Acknowledgments}

This work was supported in part by New York State through the Goergen Institute for Data Science at the University of Rochester. We thank VisualDX for discussions related to this work. 


\bibliography{references}

\begin{thebibliography}{}

\bibitem[\protect\citeauthoryear{Arevalo \bgroup et al\mbox.\egroup
  }{2015}]{arevalo2015unsupervised}
Arevalo, J.; Cruz-Roa, A.; Arias, V.; Romero, E.; and Gonz{\'a}lez, F.~A.
\newblock 2015.
\newblock An unsupervised feature learning framework for basal cell carcinoma
  image analysis.
\newblock {\em Artificial intelligence in medicine}.

\bibitem[\protect\citeauthoryear{Arroyo and
  Zapirain}{2014}]{arroyo2014automated}
Arroyo, J., and Zapirain, B.
\newblock 2014.
\newblock Automated detection of melanoma in dermoscopic images.
\newblock In Scharcanski, J., and Celebi, M.~E., eds., {\em Computer Vision
  Techniques for the Diagnosis of Skin Cancer}, Series in BioEngineering.
  Springer Berlin Heidelberg.
\newblock  139--192.

\bibitem[\protect\citeauthoryear{Caruana}{1997}]{DBLP:journals/ml/Caruana97}
Caruana, R.
\newblock 1997.
\newblock Multitask learning.
\newblock {\em Machine Learning} 28(1):41--75.

\bibitem[\protect\citeauthoryear{Cecil, Goldman, and
  Schafer}{2012}]{cecil2012goldman}
Cecil, R.~L.; Goldman, L.; and Schafer, A.~I.
\newblock 2012.
\newblock {\em Goldman's Cecil Medicine}.
\newblock Philadephia: Elsevier/Saunders, 23th edition.

\bibitem[\protect\citeauthoryear{Cox and Coulson}{2004}]{cox2004diagnosis}
Cox, N., and Coulson, I.
\newblock 2004.
\newblock Diagnosis of skin disease.
\newblock {\em Rook's Textbook of Dermatology, 7th edn. Oxford: Blackwell
  Science} 5.

\bibitem[\protect\citeauthoryear{Cruz-Roa \bgroup et al\mbox.\egroup
  }{2014}]{cruz2014automatic}
Cruz-Roa, A.; Basavanhally, A.; Gonz{\'a}lez, F.; Gilmore, H.; Feldman, M.;
  Ganesan, S.; Shih, N.; Tomaszewski, J.; and Madabhushi, A.
\newblock 2014.
\newblock Automatic detection of invasive ductal carcinoma in whole slide
  images with convolutional neural networks.
\newblock In {\em SPIE Medical Imaging},  904103--904103.
\newblock International Society for Optics and Photonics.

\bibitem[\protect\citeauthoryear{Deng \bgroup et al\mbox.\egroup
  }{2009}]{DBLP:conf/cvpr/DengDSLL009}
Deng, J.; Dong, W.; Socher, R.; Li, L.; Li, K.; and Li, F.
\newblock 2009.
\newblock Imagenet: {A} large-scale hierarchical image database.
\newblock In {\em 2009 {IEEE} Computer Society Conference on Computer Vision
  and Pattern Recognition {(CVPR} 2009), 20-25 June 2009, Miami, Florida,
  {USA}},  248--255.

\bibitem[\protect\citeauthoryear{Donahue \bgroup et al\mbox.\egroup
  }{2014}]{DBLP:conf/icml/DonahueJVHZTD14}
Donahue, J.; Jia, Y.; Vinyals, O.; Hoffman, J.; Zhang, N.; Tzeng, E.; and
  Darrell, T.
\newblock 2014.
\newblock Decaf: {A} deep convolutional activation feature for generic visual
  recognition.
\newblock In {\em Proceedings of the 31th International Conference on Machine
  Learning, {ICML} 2014, Beijing, China, 21-26 June 2014},  647--655.

\bibitem[\protect\citeauthoryear{Esteva, Kuprel, and
  Thrun}{2015}]{esteva2015deep}
Esteva, A.; Kuprel, B.; and Thrun, S.
\newblock 2015.
\newblock Deep networks for early stage skin disease and skin cancer
  classification.

\bibitem[\protect\citeauthoryear{Everingham \bgroup et al\mbox.\egroup
  }{2010}]{DBLP:journals/ijcv/EveringhamGWWZ10}
Everingham, M.; Gool, L. J.~V.; Williams, C. K.~I.; Winn, J.~M.; and Zisserman,
  A.
\newblock 2010.
\newblock The pascal visual object classes {(VOC)} challenge.
\newblock {\em International Journal of Computer Vision} 88(2):303--338.

\bibitem[\protect\citeauthoryear{Fabbrocini \bgroup et al\mbox.\egroup
  }{2014}]{Fabbrocini:2014yq}
Fabbrocini, G.; Vita, V.; Cacciapuoti, S.; Leo, G.; Liguori, C.; Paolillo, A.;
  Pietrosanto, A.; and Sommella, P.
\newblock 2014.
\newblock Automatic diagnosis of melanoma based on the 7-point checklist.
\newblock In Scharcanski, J., and Celebi, M.~E., eds., {\em Computer Vision
  Techniques for the Diagnosis of Skin Cancer}, Series in BioEngineering.
  Springer Berlin Heidelberg.
\newblock  71--107.

\bibitem[\protect\citeauthoryear{Hand and
  Chellappa}{2016}]{DBLP:journals/corr/HandC16}
Hand, E.~M., and Chellappa, R.
\newblock 2016.
\newblock Attributes for improved attributes: {A} multi-task network for
  attribute classification.
\newblock {\em CoRR} abs/1604.07360.

\bibitem[\protect\citeauthoryear{He \bgroup et al\mbox.\egroup
  }{2015}]{DBLP:journals/corr/HeZRS15}
He, K.; Zhang, X.; Ren, S.; and Sun, J.
\newblock 2015.
\newblock Deep residual learning for image recognition.
\newblock {\em CoRR} abs/1512.03385.

\bibitem[\protect\citeauthoryear{Jia \bgroup et al\mbox.\egroup
  }{2014}]{DBLP:conf/mm/JiaSDKLGGD14}
Jia, Y.; Shelhamer, E.; Donahue, J.; Karayev, S.; Long, J.; Girshick, R.~B.;
  Guadarrama, S.; and Darrell, T.
\newblock 2014.
\newblock Caffe: Convolutional architecture for fast feature embedding.
\newblock In {\em Proceedings of the {ACM} International Conference on
  Multimedia, {MM} '14, Orlando, FL, USA, November 03 - 07, 2014},  675--678.

\bibitem[\protect\citeauthoryear{Kawahara, BenTaieb, and
  Hamarneh}{2016}]{DBLP:conf/isbi/KawaharaBH16}
Kawahara, J.; BenTaieb, A.; and Hamarneh, G.
\newblock 2016.
\newblock Deep features to classify skin lesions.
\newblock In {\em 13th {IEEE} International Symposium on Biomedical Imaging,
  {ISBI} 2016, Prague, Czech Republic, April 13-16, 2016},  1397--1400.

\bibitem[\protect\citeauthoryear{Krizhevsky, Sutskever, and
  Hinton}{2012}]{DBLP:conf/nips/KrizhevskySH12}
Krizhevsky, A.; Sutskever, I.; and Hinton, G.~E.
\newblock 2012.
\newblock Imagenet classification with deep convolutional neural networks.
\newblock In {\em Advances in Neural Information Processing Systems 25: 26th
  Annual Conference on Neural Information Processing Systems 2012. Proceedings
  of a meeting held December 3-6, 2012, Lake Tahoe, Nevada, United States.},
  1106--1114.

\bibitem[\protect\citeauthoryear{Lawrence and Cox}{2002}]{lawrence2002physical}
Lawrence, C.~M., and Cox, N.~H.
\newblock 2002.
\newblock {\em Physical Signs in Dermatology}.
\newblock London: Mosby, 2nd edition.

\bibitem[\protect\citeauthoryear{Liao, Li, and Luo}{2016}]{liao2016skin}
Liao, H.; Li, Y.; and Luo, J.
\newblock 2016.
\newblock Skin disease classification versus skin lesion characterization:
  Achieving robust diagnosis using multi-label deep neural networks.
\newblock In {\em International Conference on Pattern Recognition (ICPR)}.

\bibitem[\protect\citeauthoryear{Liao}{}]{liao2016deep}
Liao, H.
\newblock A deep learning approach to universal skin disease classification.

\bibitem[\protect\citeauthoryear{Lin \bgroup et al\mbox.\egroup
  }{2014}]{DBLP:conf/eccv/LinMBHPRDZ14}
Lin, T.; Maire, M.; Belongie, S.~J.; Hays, J.; Perona, P.; Ramanan, D.;
  Doll{\'{a}}r, P.; and Zitnick, C.~L.
\newblock 2014.
\newblock Microsoft {COCO:} common objects in context.
\newblock In {\em Computer Vision - {ECCV} 2014 - 13th European Conference,
  Zurich, Switzerland, September 6-12, 2014, Proceedings, Part {V}},  740--755.

\bibitem[\protect\citeauthoryear{Ranjan, Patel, and
  Chellappa}{2016}]{DBLP:journals/corr/RanjanPC16}
Ranjan, R.; Patel, V.~M.; and Chellappa, R.
\newblock 2016.
\newblock Hyperface: {A} deep multi-task learning framework for face detection,
  landmark localization, pose estimation, and gender recognition.
\newblock {\em CoRR} abs/1603.01249.

\bibitem[\protect\citeauthoryear{Razavian \bgroup et al\mbox.\egroup
  }{2014}]{DBLP:conf/cvpr/RazavianASC14}
Razavian, A.~S.; Azizpour, H.; Sullivan, J.; and Carlsson, S.
\newblock 2014.
\newblock {CNN} features off-the-shelf: An astounding baseline for recognition.
\newblock In {\em {IEEE} Conference on Computer Vision and Pattern Recognition,
  {CVPR} Workshops 2014, Columbus, OH, USA, June 23-28, 2014},  512--519.

\bibitem[\protect\citeauthoryear{Ren \bgroup et al\mbox.\egroup
  }{2015}]{DBLP:conf/nips/RenHGS15}
Ren, S.; He, K.; Girshick, R.~B.; and Sun, J.
\newblock 2015.
\newblock Faster {R-CNN:} towards real-time object detection with region
  proposal networks.
\newblock In {\em Advances in Neural Information Processing Systems 28: Annual
  Conference on Neural Information Processing Systems 2015, December 7-12,
  2015, Montreal, Quebec, Canada},  91--99.

\bibitem[\protect\citeauthoryear{Russakovsky \bgroup et al\mbox.\egroup
  }{2015}]{DBLP:journals/ijcv/RussakovskyDSKS15}
Russakovsky, O.; Deng, J.; Su, H.; Krause, J.; Satheesh, S.; Ma, S.; Huang, Z.;
  Karpathy, A.; Khosla, A.; Bernstein, M.~S.; Berg, A.~C.; and Li, F.
\newblock 2015.
\newblock Imagenet large scale visual recognition challenge.
\newblock {\em International Journal of Computer Vision} 115(3):211--252.

\bibitem[\protect\citeauthoryear{Simonyan and
  Zisserman}{2014}]{DBLP:journals/corr/SimonyanZ14a}
Simonyan, K., and Zisserman, A.
\newblock 2014.
\newblock Very deep convolutional networks for large-scale image recognition.
\newblock {\em CoRR} abs/1409.1556.

\bibitem[\protect\citeauthoryear{Wang \bgroup et al\mbox.\egroup
  }{2014}]{wang2014cascaded}
Wang, H.; Cruz-Roa, A.; Basavanhally, A.; Gilmore, H.; Shih, N.; Feldman, M.;
  Tomaszewski, J.; Gonzalez, F.; and Madabhushi, A.
\newblock 2014.
\newblock Cascaded ensemble of convolutional neural networks and handcrafted
  features for mitosis detection.
\newblock In {\em SPIE Medical Imaging},  90410B--90410B.
\newblock International Society for Optics and Photonics.

\bibitem[\protect\citeauthoryear{Xie \bgroup et al\mbox.\egroup
  }{2014}]{xie2014dermoscopy}
Xie, F.; Wu, Y.; Jiang, Z.; and Meng, R.
\newblock 2014.
\newblock Dermoscopy image processing for chinese.
\newblock In Scharcanski, J., and Celebi, M.~E., eds., {\em Computer Vision
  Techniques for the Diagnosis of Skin Cancer}, Series in BioEngineering.
  Springer Berlin Heidelberg.
\newblock  109--137.

\bibitem[\protect\citeauthoryear{Zeiler and
  Fergus}{2014}]{DBLP:conf/eccv/ZeilerF14}
Zeiler, M.~D., and Fergus, R.
\newblock 2014.
\newblock Visualizing and understanding convolutional networks.
\newblock In {\em Computer Vision - {ECCV} 2014 - 13th European Conference,
  Zurich, Switzerland, September 6-12, 2014, Proceedings, Part {I}},  818--833.

\bibitem[\protect\citeauthoryear{Zhang \bgroup et al\mbox.\egroup
  }{2016}]{DBLP:journals/pami/ZhangLLT16}
Zhang, Z.; Luo, P.; Loy, C.~C.; and Tang, X.
\newblock 2016.
\newblock Learning deep representation for face alignment with auxiliary
  attributes.
\newblock {\em {IEEE} Trans. Pattern Anal. Mach. Intell.} 38(5):918--930.

\bibitem[\protect\citeauthoryear{Zhou \bgroup et al\mbox.\egroup
  }{2015}]{DBLP:journals/corr/ZhouKLOT15}
Zhou, B.; Khosla, A.; Lapedriza, {\`{A}}.; Oliva, A.; and Torralba, A.
\newblock 2015.
\newblock Learning deep features for discriminative localization.
\newblock {\em CoRR} abs/1512.04150.

\end{thebibliography}
\bibliographystyle{aaai}

\end{document}